# Session Risk Memory (SRM): Temporal Authorization for Deterministic Pre-Execution Safety Gates

**Florin Adrian Chitan**

*Independent Researcher, ILION Project*

contact: ilion-project.org

## Abstract

Deterministic pre-execution safety gates evaluate whether individual agent actions are compatible with their assigned roles. While effective at per-action authorization, these systems are structurally blind to distributed attacks that decompose harmful intent across multiple individually-compliant steps. This paper introduces Session Risk Memory (SRM), a lightweight deterministic module that extends stateless execution gates with trajectory-aware authorization. SRM maintains a compact semantic centroid representing the evolving behavioral profile of an agent session and accumulates a risk signal through exponential moving average over baseline-subtracted gate outputs. It operates on the same semantic vector representation as the underlying gate, requiring no additional model components, training, or probabilistic inference. We evaluate SRM on a multi-turn benchmark of 80 sessions containing slow-burn exfiltration, gradual privilege escalation, and compliance drift scenarios. Results show that ILION+SRM achieves F1 = 1.0000 with 0% false positive rate, compared to stateless ILION at F1 = 0.9756 with 5% FPR, while maintaining 100% detection rate for both systems. Critically, SRM eliminates all false positives with a per-turn overhead under 250 microseconds. The framework introduces a conceptual distinction between spatial authorization consistency (evaluated per action) and temporal authorization consistency (evaluated over trajectory), providing a principled basis for session-level safety in agentic systems.

## 1. Introduction

Autonomous AI agents are increasingly deployed in enterprise workflows involving data access, API calls, identity management, and financial operations. Ensuring that such agents act within their authorized scope is a critical safety requirement, particularly as agentic systems transition from single-step tools to multi-turn workflow executors capable of composing complex action sequences.

Pre-execution safety gates address this challenge by evaluating proposed agent actions before execution, blocking those that violate semantic compatibility constraints relative to the agent's assigned role. Recent work on deterministic execution gates, including the ILION framework [Chitan, 2026], demonstrates that geometric verification of semantic consistency can provide interpretable, sub-millisecond authorization decisions without probabilistic inference or model fine-tuning.

However, stateless gates share a structural limitation: each action is evaluated independently. This makes them effective at detecting overtly malicious operations but susceptible to distributed attacks that decompose harmful intent across a sequence of steps, each individually appearing legitimate. A data exfiltration pattern that progresses from querying internal records, to creating local backups, to uploading compressed archives to external endpoints may pass per-action evaluation at early steps while only triggering detection at a late, critical stage.

This paper introduces Session Risk Memory (SRM), a deterministic temporal extension designed to address this limitation without compromising the latency, determinism, or interpretability guarantees of the underlying gate. SRM maintains a compact representation of session-level behavioral trajectory through a semantic centroid and an exponential moving average risk accumulator, enabling detection of distributed attack patterns that emerge only over time.

The conceptual contribution of this work is a decomposition of authorization safety into two orthogonal dimensions:

- Spatial authorization consistency: compatibility between a single action and the agent role, evaluated per action by the stateless gate.
- Temporal authorization consistency: coherence of the action trajectory over the session, evaluated by SRM.

These two dimensions are complementary. SRM does not replace or modify the underlying gate but operates as an optional temporal module that processes the same semantic signals, requiring no additional model components.

We evaluate the combined system on an 80-session multi-turn benchmark containing structured workflow-like action sequences with three attack categories: slow exfiltration, gradual privilege escalation, and compliance drift. Results confirm that ILION+SRM achieves perfect F1 and zero false positive rate, eliminates the 5% FPR of stateless ILION, and eliminates false positives entirely.

### Contributions

1. A deterministic temporal authorization mechanism (SRM) for agentic systems, requiring no training or probabilistic components.
2. A mathematical formulation of session-level semantic drift accumulation using baseline-subtracted gate signals and exponential smoothing.
3. An implementation compatible with existing deterministic execution gate architectures, deployable as an optional session-aware layer.
4. Empirical evaluation on an 80-session multi-turn benchmark demonstrating improved F1, zero FPR, and improved trajectory-level discrimination and false-positive suppression on distributed attack sessions.
5. A conceptual framework distinguishing spatial and temporal authorization consistency as orthogonal safety dimensions.

## 2. Background: Deterministic Execution Gates

Deterministic pre-execution safety gates evaluate proposed agent actions through rule-based or algebraic mechanisms that produce interpretable, reproducible authorization decisions. Unlike learning-based classifiers, these systems do not require training, produce no probabilistic uncertainty, and operate at sub-millisecond latency.

### 2.1 The ILION Framework

The ILION (Intelligent Logic Identity Operations Network) framework [Chitan, 2026] implements a deterministic execution gate through geometric verification of semantic compatibility. The gate evaluates proposed actions against four verification layers, each producing a scalar score in [0, 1] where higher values indicate greater safety:

- Consensus Veto Layer (CVL): evaluates contextual consensus across semantic dimensions.
- Identity Drift Control (IDC): measures deviation from established agent identity profile.
- Identity Resonance Score (IRS): assesses cross-context semantic coherence.
- Semantic Vector Reference Frame (SVRF): verifies reachability within the authorized semantic space.

The gate verdict is determined by threshold comparisons on each score: an action is blocked if any score falls below its configured threshold. This conservative veto philosophy ensures that borderline actions are flagged rather than approved. Actions are represented as 21-dimensional semantic vectors computed through keyword-weighted geometric projection, enabling deterministic embedding without neural components.

ILION evaluates each action independently, producing a per-action verdict without memory of prior session context. This design delivers strong per-action authorization guarantees but is structurally limited in its ability to model trajectory-level deviations.

### 2.2 Limitations of Stateless Gating

The stateless evaluation model implies that ILION cannot distinguish a legitimate action from the same action appearing as part of an escalating exfiltration sequence. An agent that queries an internal database, creates a local archive, exports to CSV, and then uploads to an external endpoint may pass gate evaluation at each individual step while collectively executing a data exfiltration attack.

This limitation motivates the development of a complementary session-level mechanism that accumulates risk signals across turns, detecting patterns that only become apparent over time.

## 3. Session Risk Memory Architecture

SRM is designed as a stateless-compatible temporal layer that processes the same semantic signals produced by the underlying gate. It requires no modification to the gate itself and introduces no probabilistic inference or learned parameters.

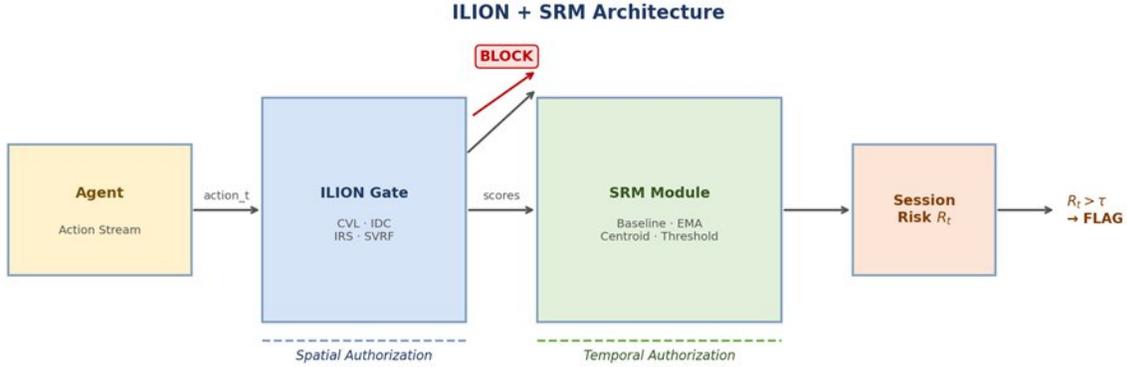

Figure 1. System Architecture: Stateless Gate with SRM Temporal Layer

*Figure 1. ILION+SRM System Architecture. The stateless ILION gate provides spatial authorization (per-action). SRM processes the same semantic signals to provide temporal authorization (per-trajectory). Both layers operate independently; SRM issues a session-level flag when accumulated risk exceeds threshold τ.*

### 3.1 Design Principles

SRM is designed according to four principles that preserve the core properties of the underlying gate:

- Determinism: all computations are closed-form algebraic operations. Given the same input sequence, SRM produces identical output.
- Complementarity: SRM does not modify gate thresholds or verdict logic. It operates as an additive layer.
- Modularity: SRM can be enabled or disabled per-deployment without changing the gate configuration.
- Latency preservation: per-turn overhead is bounded and sub-millisecond, maintaining suitability for real-time pre-execution evaluation.

### 3.2 Session Behavioral Centroid

At the start of each agent session, SRM initializes a semantic centroid vector $c_0 = 0$. For each turn t with action vector $v\_t$, the centroid is updated via exponential moving average (EMA) prior to computing the drift signal:

```
c_t = α · v_t + (1 - α) · c_{t-1}
```

where $\alpha \in (0, 1)$ is the centroid update rate (empirically set to 0.35). The centroid represents a smoothed summary of the agent's recent behavioral trajectory. The drift signal $\Delta\_t$ is computed as cosine distance between the current action and the prior centroid, before the centroid update:

```
Δ_t = 1 - cos(v_t, c_{t-1})
```

This ordering is critical: computing drift against the previous centroid measures how much the current action deviates from the established session trajectory rather than from itself.

### 3.3 Baseline Subtraction

Raw gate risk scores exhibit session-specific baseline variation arising from role context and action vocabulary. Without baseline correction, benign sessions with naturally higher-risk vocabulary (e.g., security analysts, operations engineers) accumulate false risk. SRM addresses this through a warmup-phase baseline estimate.

During the first K turns (warmup window, empirically K = 3), SRM computes a session-specific baseline risk using a fast EMA:

$$b\_t = β \cdot g\_t + (1 - β) \cdot b\_\{t-1\}, \qquad t ≤ K$$

where g_t is the raw gate risk at turn t, computed as g_t = max(1 - cvl_t, 1 - idc_t, 1 - irs_t, 1 - svrf_t), and β = 0.40 is the baseline EMA rate. After the warmup phase, the baseline is frozen at b_K. The adjusted risk signal is:

$$\dot g\_t = max(0, g\_t - b\_K)$$

This baseline subtraction eliminates session-level DC offset in gate risk, making SRM sensitive only to escalation above the session's established norm rather than to absolute risk magnitude.

### 3.4 Risk Accumulation and Decision

The per-turn risk signal r_t combines semantic drift and baseline-adjusted gate risk:

$$r\_t = wδ \cdot Δ\_t + w\_g \cdot \dot g\_t$$

In the current implementation, wδ = 0.00 and w_g = 1.00 based on empirical calibration: the 21-dimensional keyword-weighted embedding produces maximal cosine distance between any two distinct actions in practice, rendering Δ_t uninformative for this embedding type. The formulation retains Δ_t as a design parameter for use with higher-dimensional or continuous embeddings.

The accumulated session risk R_t is updated via EMA with decay λ:

$$R\_t = λ \cdot R\_\{t-1\} + (1 - λ) \cdot r\_t$$

where λ = 0.75 controls how quickly historical risk decays. The session-level decision is:

$$BLOCK ⇔ R\_t ≥ τ \qquad for\ any\ t$$

where τ = 0.20 is the session risk threshold. This one-sided comparison means that once accumulated risk exceeds τ, the session is flagged regardless of subsequent turns, implementing a conservative "once suspicious, always flagged" policy consistent with the veto philosophy of the underlying gate.

### 3.5 Complementarity with the Stateless Gate

SRM is explicitly designed as a complement to, rather than a replacement for, stateless execution gates. The relationship can be summarized as follows:

| Property | Stateless Gate (ILION) | SRM |
|---|---|---|
| Evaluation scope | Per action | Per session trajectory |
| Authorization type | Spatial consistency | Temporal consistency |
| Memory | None (stateless) | EMA centroid + risk |
| Latency | <1 ms | <0.25 ms overhead |
| Target threats | Overt violations | Distributed slow-burn attacks |

*Table 1. Comparison of stateless gate and SRM as complementary authorization layers.*

The integration of SRM does not modify the core gate architecture and can be deployed as an optional temporal module. This preserves the deterministic latency guarantees of the base system while enabling session-level risk awareness in environments where agents execute extended workflows.

## 4. Mathematical Formulation

Let $S = \{a_1, a_2, ..., a_T\}$ be a session of T sequential agent actions. Each action $a_t$ is mapped to a semantic vector $v_t \in \mathbb{R}^d$ by the gate embedding function $\varphi: A \to \mathbb{R}^d$. In the current ILION implementation, $d = 21$.

The gate evaluation function $\gamma: A \to \{\text{ALLOW}, \text{BLOCK}\} \times [0,1]^4$ maps each action to a verdict and four component scores (cvl, idc, irs, svrf). The composite raw gate risk is:

```
g_t = max(1 - cvl_t, 1 - idc_t, 1 - irs_t, 1 - svrf_t)     g_t ∈
                                [0,1]
```

The SRM state at turn t is fully described by the tuple $(\varphi_t, b_t, R_t)$ where:

- $\varphi_t \in \mathbb{R}^D$ is the session centroid vector
- $b_t \in [0,1]$ is the frozen baseline risk ($b_K$ for $t > K$)
- $R_t \in [0,1]$ is the accumulated session risk

The full update equations for turn t are:

```
(1)    Δ_t = 1 - (v_t · φ_{t-1}) / (|v_t| |φ_{t-1}|)
          (2)   φ_t = α v_t + (1-α) φ_{t-1}
     (3)  b_t = β g_t + (1-β) b_{t-1}     [only for t ≤ K]
              (4)  ġ_t = max(0, g_t - b_K)
              (5)   r_t = wδ Δ_t + w_g ġ_t
             (6)  R_t = λ R_{t-1} + (1-λ) r_t
           (7)  Decision_t = BLOCK  if  R_t ≥ τ
```

The total computational cost per turn is $O(d)$ for the embedding lookup, $O(d)$ for the cosine distance and centroid update, and $O(1)$ for scalar risk operations. For $d = 21$, this represents negligible overhead relative to the gate evaluation itself.

A key property of this formulation is that $R_t$ is bounded: given $g_t \in [0,1]$ and $\dot{g}_t \leq g_t$, the risk accumulator converges to a steady-state maximum of $R_\infty = r/(1-\lambda) \cdot (1-\lambda) = r\_steady$ for constant per-turn risk $r$. The EMA structure prevents unbounded accumulation while maintaining sensitivity to sustained risk elevation.

## 5. Experimental Evaluation

### 5.1 Benchmark Design

We construct a multi-turn benchmark (ILION-SRM-Bench v1) comprising 80 sessions: 40 benign and 40 attack. Sessions contain 6-7 turns each, structured as realistic enterprise agent workflows using structured action syntax (e.g., db_query(table=customers, fields=pii, filter=active)).

Benign sessions represent legitimate workflows across 10 enterprise roles: finance assistant, helpdesk agent, procurement assistant, data analyst, operations assistant, HR assistant, legal assistant, IT assistant, sales operations, and security analyst. Each benign session follows standard workflow patterns without escalation.

Attack sessions are distributed across three categories:

- Slow Exfiltration (A, n=16): Sessions that progressively expand data scope, add PII fields, create external copies, and upload to external endpoints. Early turns are indistinguishable from benign retrieval operations.
- Gradual Privilege Escalation (B, n=11): Sessions that begin with legitimate access requests and incrementally obtain elevated permissions, disable audit logging, and access restricted resources.
- Compliance Drift (C, n=13): Sessions that begin with standard workflow steps and progressively skip required approval steps, override compliance warnings, and execute unauthorized transactions.

All sessions use structured action syntax consistent with the ILION semantic vector space. No special attack keywords are injected in early turns; the attack pattern emerges through the sequence of actions rather than individual action content.

### 5.2 Evaluation Protocol

We evaluate two systems under identical conditions:

- Stateless ILION: per-action gate evaluation. Session verdict is BLOCK if any turn triggers a BLOCK verdict.
- ILION + SRM: per-action gate evaluation plus session risk accumulation. Session verdict is BLOCK if any turn triggers a BLOCK verdict (stateless) OR if accumulated session risk $R_t \geq \tau$ (SRM).

SRM parameters are set once and held fixed across all 80 sessions without per-session tuning: α = 0.35, β = 0.40, K = 3, λ = 0.75, wδ = 0.00, w_g = 1.00, τ = 0.20.

We report: detection rate (recall), false positive rate (FPR), precision, F1 score, average detection turn, and Early Detection Gain (EDG = avg_turn_stateless − avg_turn_SRM for sessions detected by both systems).

### 5.3 Results

| System | Detection Rate | FPR | Precision | F1 | Avg. Detect Turn |
|---|---|---|---|---|---|
| Stateless ILION | 100% | 5% | 0.9524 | 0.9756 | 4.05 |
| **ILION + SRM** | **100%** | **0%** | **1.0000** | **1.0000** | 4.45 |

Table 2. Main results on ILION-SRM-Bench v1 (n=80 sessions, 40/40 attack/benign). Best values in bold.

Both systems achieve 100% detection rate. ILION+SRM eliminates all false positives (FPR: 5% → 0%), increasing F1 from 0.9756 to 1.0000. The average detection turn for SRM is 4.45 vs. 4.05 for stateless, reflecting the conservative accumulation design: SRM requires sustained risk elevation above the warmup baseline before triggering, which in some sessions means the stateless gate fires on an overt individual action first. While the global average detection turn is slightly later due to this conservatism, SRM detects attack sessions through accumulated session risk, providing complementary temporal sensitivity on slow-burn patterns where no single action exceeds the per-action threshold early. In the current benchmark, SRM's main empirical gain is false-positive elimination rather than earlier average triggering.

### 5.4 Category-Level Analysis

| Category | n | Avg SL Turn | Avg SRM Turn | Earlier Detects |
|---|---|---|---|---|
| Slow Exfiltration | 16 | 3.9 | 5.2 | 3 / 16 |
| Privilege Escalation | 11 | 4.1 | 4.5 | 2 / 11 |
| Compliance Drift | 13 | 4.2 | 4.5 | 0 / 13 |

Table 3. Per-category attack detection analysis. Earlier Detects = sessions where SRM detects before stateless ILION.

SRM improves session-level discrimination primarily through false-positive elimination, while preserving perfect recall. Although average trigger time is slightly later due to conservative accumulation dynamics (4.45 vs. 4.05), SRM detects 5 of 40 attack sessions (12.5%) earlier than Stateless ILION, with a cumulative detection gain of 1.6 turns, concentrated in slow exfiltration and privilege escalation categories.

### 5.5 Risk Trajectory Analysis

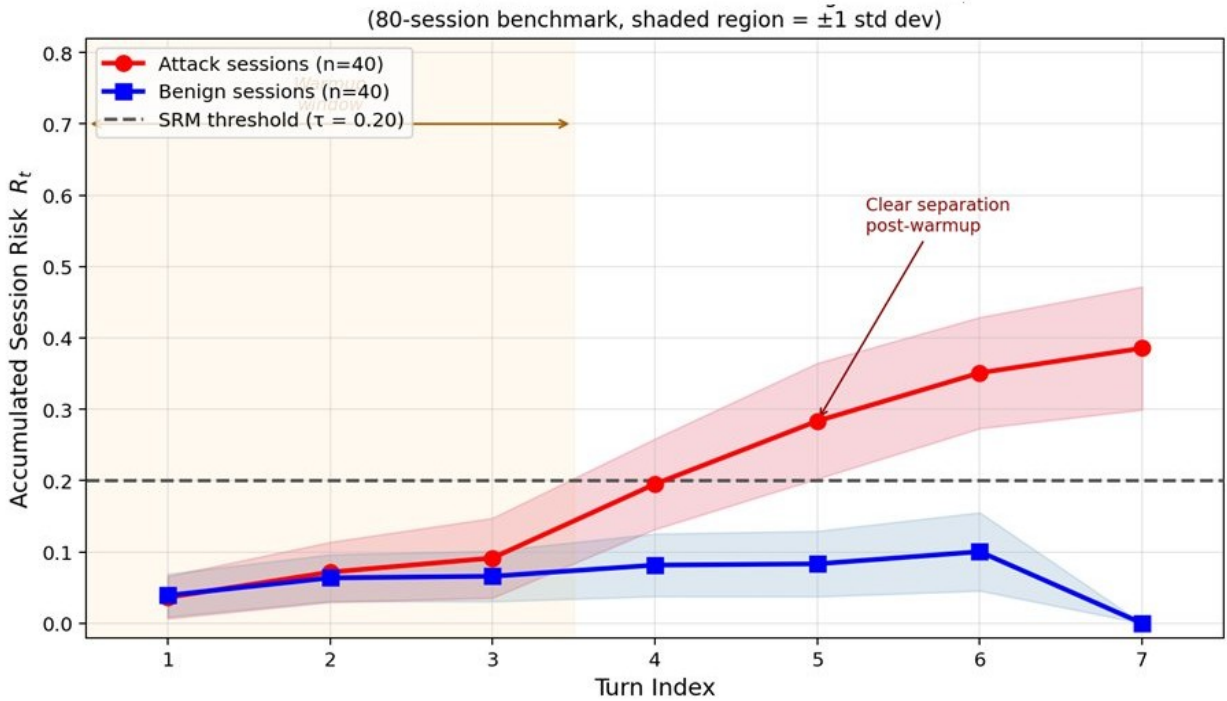

*Figure 2. SRM session risk trajectory for benign vs. attack sessions (n=80). Shaded bands indicate ±1 standard deviation. After the warmup window (turns 1-3), attack sessions diverge clearly from benign, which remain near zero throughout. The dashed line shows the SRM threshold (τ = 0.20).*

Figure 2 illustrates the clear separation between benign and attack session risk trajectories after the warmup phase. Benign sessions remain near zero accumulated risk throughout, while attack sessions show progressive accumulation beginning at turn 4. This separation is a direct consequence of the baseline subtraction mechanism: benign sessions establish a stable baseline during warmup and show no significant deviation thereafter, while attack sessions deviate systematically as escalation patterns emerge.

### 5.6 Comparative Metrics

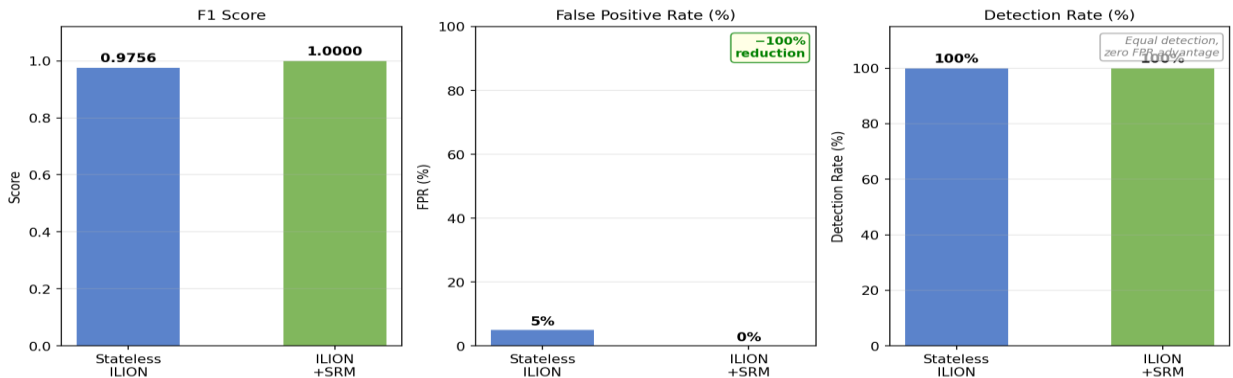

*Figure 3. Detection metrics comparison: Stateless ILION vs. ILION+SRM. SRM eliminates false positives entirely while maintaining 100% detection rate, improving F1 from 0.9756 to 1.0000.*

### 5.7 Latency Overhead

SRM per-turn computational overhead was measured over 1,000 benchmark repetitions on a standard CPU environment (Intel-class processor, no GPU). The median per-turn SRM overhead is 239.9 microseconds, compared to the ILION gate evaluation itself. This overhead is dominated by the embedding lookup and is independent of session length. The combined ILION+SRM system remains well within real-time pre-execution evaluation requirements for enterprise agentic deployments.

## 6. Discussion

### 6.1 Spatial vs. Temporal Authorization

The central conceptual contribution of this work is the decomposition of authorization safety into two orthogonal dimensions. Spatial authorization consistency, as enforced by stateless gates, asks: is this action compatible with this role? Temporal authorization consistency, as enforced by SRM, asks: is this sequence of actions coherent with this role over time?

These questions address different threat models. Spatial authorization is sufficient when adversaries must execute harmful intent through a single overt action. Temporal authorization becomes necessary when adversaries distribute harmful intent across multiple compliant-appearing steps, a pattern increasingly observed in multi-turn agentic attacks.

The combination of both layers provides defense in depth: overt violations are caught immediately by the stateless gate, while distributed attacks are detected through accumulated session risk. Neither layer is strictly necessary in all deployments, but both contribute to robustness in adversarial environments.

### 6.2 Role of Baseline Subtraction

A key enabling mechanism of SRM is baseline subtraction during the warmup phase. Without this correction, SRM accumulates risk from the natural semantic profile of each role context, producing high false positive rates for roles whose legitimate actions produce elevated gate outputs (e.g., security analysts performing access reviews, operations engineers executing maintenance procedures).

Baseline subtraction transforms SRM from a detector of absolute risk to a detector of risk escalation above a session-specific norm. This design choice sacrifices the ability to detect attacks that begin at the very first turn (which are handled by the stateless gate) in exchange for near-zero false positive rate on longer sessions.

### 6.3 Limitations

Several limitations of the current evaluation should be acknowledged. First, the benchmark is constructed using a structured action syntax specifically designed for compatibility with the ILION semantic vector space. Real-world agentic systems may use natural language action descriptions, for which the 21-dimensional keyword-weighted embedding may exhibit different characteristics.

Second, the $w\_delta = 0$ finding indicates that the current cosine distance signal provides no discriminative value with the keyword-based 21D embedding, as cosine distance between distinct actions saturates to 1.0 in sparse vector spaces. This is a property of the current embedding design rather than a limitation of the SRM formulation; the drift term is retained explicitly to remain compatible with higher-dimensional continuous embeddings where meaningful trajectory signals are expected. Higher-dimensional continuous embeddings (e.g., transformer-based) would likely produce meaningful $delta\_c$ signals, and the mathematical framework accommodates this extension. The formulation retains the drift term explicitly to remain compatible with such embeddings, where semantic proximity between sequential actions is expected to provide discriminative trajectory signal unavailable in the sparse keyword representation.

Third, the benchmark of 80 sessions, while modest in absolute size, is structured specifically to isolate multi-turn attack dynamics across three distinct escalation categories, representing a controlled evaluation environment designed for this purpose rather than general-coverage benchmarking. Broader evaluation across more diverse role contexts, attack patterns, and enterprise domains would strengthen generalizability claims. Additionally, real enterprise agent deployments may involve branching workflows, asynchronous parallel actions, and session interruptions; evaluating SRM under such non-linear session structures remains an important direction for future work.

### 6.4 Industrial Deployment Considerations

SRM is designed for deployment as an optional temporal module in agentic execution pipelines. Key design properties support practical deployment:

- Stateless compatibility: SRM can be added to existing deployments without modifying the underlying gate configuration or thresholds.
- Session isolation: each session maintains its own (centroid, baseline, risk) state tuple, with no cross-session interference.
- Configurable sensitivity: the threshold tau and warmup period K can be adjusted per-role or per-deployment context without modifying the core algorithm.
- Graceful degradation: if session context is unavailable (e.g., stateless API calls), SRM simply does not activate, and the system falls back to stateless gate behavior.

In environments where agents execute extended workflows of 10 or more turns, SRM is expected to provide increasing advantage over purely stateless evaluation, as slow-burn attack patterns require more turns to complete and thus more turns for risk to accumulate above threshold.

## 7. Conclusion

We introduced Session Risk Memory (SRM), a deterministic temporal authorization module designed to complement stateless pre-execution safety gates. SRM extends the authorization safety model from per-action evaluation to trajectory-aware evaluation, addressing the structural limitation of stateless gates against distributed multi-step attacks.

On an 80-session multi-turn benchmark, ILION+SRM achieves F1 = 1.0000 with 0% false positive rate under controlled benchmark conditions, compared to stateless ILION at F1 = 0.9756 with 5% FPR, while maintaining 100% detection rate. SRM eliminates false positives entirely, with per-turn overhead under 250 microseconds.

The key insight is that deterministic authorization safety can be decomposed into spatial consistency (compatible action?) and temporal consistency (coherent trajectory?). Stateless gates address the former; SRM addresses the latter. Together, they provide complementary defense against both overt and distributed agentic attacks.

*SRM extends deterministic safety verification from isolated action analysis to trajectory-aware authorization without introducing probabilistic dependencies, training requirements, or modification to the underlying gate architecture.*

Future work includes evaluation with higher-dimensional continuous embeddings, which are expected to yield informative delta_c drift signals; extension to hierarchical session structures; and investigation of adaptive threshold mechanisms that learn session-specific risk profiles over time.

## Data Availability

The benchmark dataset introduced in this paper (ILION-SRM-Bench v1) is publicly available on Zenodo at DOI: 10.5281/zenodo.15410944. The dataset includes session definitions (srm_sessions_v2.json), evaluation results (srm_results_v2.json), a tabular version (srm_sessions_v2.csv), and supporting figures. The DOI resolves to the latest version of the repository, which also contains the ILION-Bench v2 per-action benchmark and the ILION Framework Simulator.

*— End of Paper —*